\documentclass[conference]{IEEEtran}
\newif\ifcameraready
\camerareadytrue

\usepackage{booktabs}
\usepackage{multirow}
\usepackage{amsmath}
\usepackage{graphicx}
\usepackage[caption=false,font=footnotesize]{subfig}
\usepackage{float}
\usepackage[numbers,sort&compress]{natbib}
\usepackage{url}
\usepackage{hyperref}
\hypersetup{
    colorlinks   = true,
    citecolor    = blue,
    urlcolor    =  blue
}

\graphicspath{{figures/}}

\newcommand{\hf}[2]{\raisebox{-2.2pt}{\includegraphics[scale=0.09]{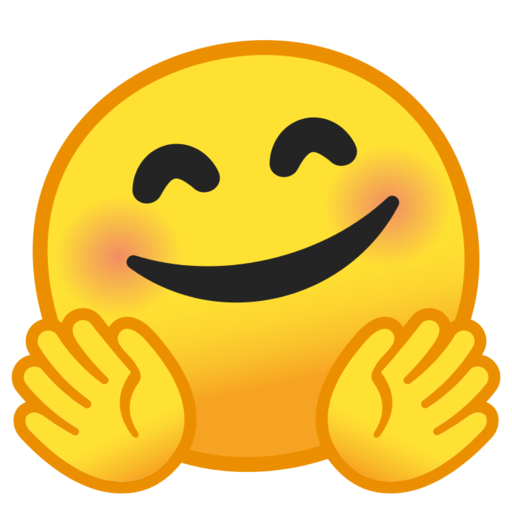}}~\href{#1}{\texttt{#2}}}

\newcommand{\gh}[2]{\raisebox{-2.2pt}{\includegraphics[scale=0.02]{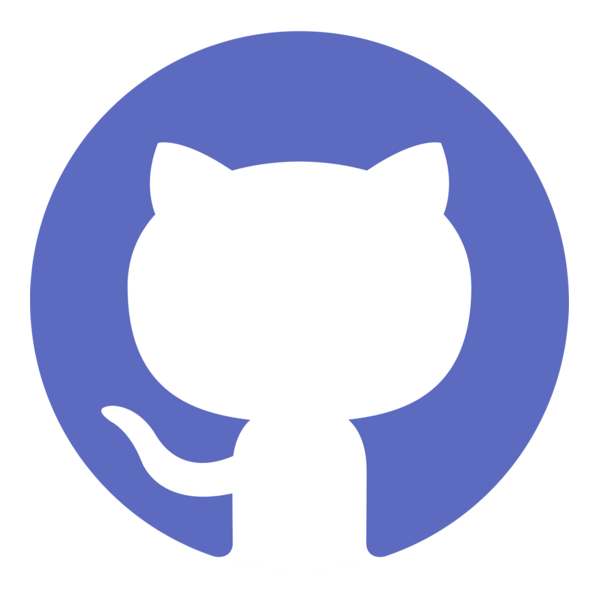}}~\href{#1}{\texttt{#2}}}

\begin{document}

\title{When Prices Double in a Week: Forecasting of Agricultural Volatility in Import-Isolated Markets}

% Double-blind: author block omitted for initial submission
\ifcameraready
\author{
\IEEEauthorblockN{%
Ranuga Weerasekara,
Heshan Nethmina,
Manuja Ranathunga,
Vinma Wettasinghe,
Dinithi Navodya,\\
Subavarshana Arumugam,
Nirasha Munasinghe,
Nisansa de Silva,
Sandareka Wickramanayake}
\IEEEauthorblockA{Dept.\ of Computer Science \& Engineering, University of Moratuwa, Sri Lanka.\\
\texttt{\{ranugaw.23, heshann.23, vinmaw.23, dinithin.23, manujar.23,}\\
\texttt{subavarshanaa.21, nirasha.25, NisansaDds, sandarekaw\}@cse.mrt.ac.lk}
}
}  
\fi

\maketitle

%% ================================================================
%%  ABSTRACT
%% ================================================================
\begin{abstract}
Vegetable prices in Sri Lanka are highly volatile because the market is largely import-isolated, so supply disruptions quickly drive prices up. This study develops a machine learning framework to forecast such volatility by incorporating supply-chain-aware features and explicitly modelling the country's two cultivation seasons, Maha (October–April) and Yala (May–September). An integrated dataset was constructed by combining retail and farmer-gate prices with origin-aligned weather variables, diesel costs, and exchange rates across 12 vegetable varieties and 14 market centres from 2013 to 2019. A gradient-boosted ensemble model (XGBoost and LightGBM) was trained and optimised using Optuna, and unified and season-specific configurations were compared. Results show that season-specific models improve within-season fit, with the Yala-specific model achieving the highest $R^{2}$ of 0.9420 (95\% CI [0.690, 1.000]), while the unified model delivers the best overall predictive accuracy of 90.84\% (95\% CI [88.34\%, 91.52\%]) and an $R^{2}$ of 0.9281 (95\% CI [0.760, 1.000]). Notably, the unified model maintains 85.96\% accuracy on a completely unseen 2024 hyperinflationary period without retraining, successfully tracking major price surges. These findings suggest that agricultural price movements in import-constrained markets are meaningfully predictable when models capture supply-chain dynamics, offering practical value for early warning and decision making by farmers, traders, and policymakers. Existing studies on Sri Lankan vegetable prices are confined to Autoregressive Integrated Moving Average (ARIMA) and Generalized Autoregressive Conditional Heteroskedasticity (GARCH) applied to single markets, with no supply-chain features, seasonal segmentation, or cross-regime validation.
\end{abstract}

\begin{IEEEkeywords}
agricultural price forecasting, XGBoost, LightGBM, import-isolated markets, 
seasonal modelling, supply-chain feature engineering, cross-regime generalisation, 
Sri Lanka
\end{IEEEkeywords}

%% ================================================================
%%  I. INTRODUCTION
%% ================================================================
\section{Introduction}\label{sec:intro}

%% STATE OF THE ART
Predicting agricultural prices is a well-studied problem. Statistical
models such as ARIMA and  Seasonal ARIMA (SARIMA) ~\cite{champika2023analysis} and Prophet~\cite{taylor2018forecasting}
have been widely used, while gradient-boosted methods like
XGBoost~\cite{chen2016xgboost} and LightGBM~\cite{ke2017lightgbm} have outperformed
them on complex, non-linear price data~\cite{madubhashini2023predicting}. However,
three problems remain unsolved across the literature: existing models
source weather data from retail markets rather than actual growing zones;
they treat a full year of prices as one continuous series, ignoring
seasonal structure; and they rarely test whether a trained model holds
up under a completely different economic regime~\cite{little2019statistical, komarek2020review}.

Sri Lanka makes all three problems worse. Its vegetable market is fully
import-isolated; every flood, drought, or supply breakdown hits retail
prices directly with no buffer, causing prices to double within a single
week~\cite{HectorKo2024Weekly}. Its two cultivation seasons, \textbf{Maha}
(October--April) and \textbf{Yala} (May--September), governed by
opposing monsoons, create structurally different supply conditions that
a single annual model cannot separate. Production further divides into
\emph{upcountry} crops (carrots, leeks, green beans) grown in the
central highlands and \emph{lowcountry} crops (brinjals, pumpkin) grown
across the wet and dry zones. Prior work on this market has relied
exclusively on ARIMA and GARCH applied to single markets and limited
varieties~\cite{champika2023analysis, madubhashini2023predicting, engle2001garch}, with no supply-chain features, seasonal segmentation, or cross-regime validation ever
attempted.

%% OBJECTIVES
This study addresses the gap through four sequential steps:
\begin{enumerate}
  \item Construct the first integrated Sri Lankan vegetable price dataset,
    resolving spatial weather mismatch and missingness through four
    novel preprocessing strategies across 12 vegetable varieties and
    14 market centres (2013--2019).
  \item Engineer supply-chain-aware features, including origin-zone
    weather lags, diesel-driven logistics costs, USD/LKR exchange rates,
    and farmer-gate spreads, to encode structural market mechanics.
  \item Train and compare unified and season-specific gradient-boosted
    ensemble models (XGBoost + LightGBM, Optuna tuned) across Maha and
    Yala seasons to quantify the precision-versus-momentum trade-off.
  \item Evaluate cross-regime generalisation on the fully unseen 2024
    hyperinflationary period without retraining, to test whether
    structural mechanics transfer across economic regimes.
\end{enumerate}

The study is guided by the following five research questions:
\begin{enumerate}
  \item \textbf{RQ1:} Are price shocks in a fully import-isolated market
    structurally predictable, or do they behave as random walks?
  \item \textbf{RQ2:} What forces govern the farmer-gate-to-retail spread
    across seasons?
  \item \textbf{RQ3:} Do Yala/Maha-specific models outperform a unified
    model, or does cross-season blindness cancel the gains?
  \item \textbf{RQ4:} Do macroeconomic step-changes (diesel, USD/LKR)
    outweigh weather signals in predictive importance?
  \item \textbf{RQ5:} Can a 2013--2019 model generalise to the 2024
    hyperinflationary regime without retraining?
\end{enumerate}

This paper is structured as follows. Section~II reviews
related work in agricultural price forecasting. Section~III describes
the dataset construction process. Section~IV presents the methodology
and modelling framework. Section~V reports experimental results, and
Section~VI discusses key insights. Finally, Section~VII concludes the
paper and outlines directions for future work. 
\hf{https://huggingface.co/datasets/teamSynapse/sri-lanka-weekly-vegetable-retail-farmer-price-dataset}{Data} and \gh{https://github.com/RanugaVW/SL-Vegetable-Price-Forecaster}{code} for this work are publicly available.

%% ================================================================
%%  II. RELATED WORK
%% ================================================================
\section{Related Work}\label{sec:related}

\subsection{Statistical Time-Series Models}

Statistical time-series models remain the dominant baseline in
agricultural price forecasting. ARIMA and its seasonal extension, SARIMA,
have been widely applied to vegetable and commodity markets in South Asia,
with Champika and Mugera~\cite{champika2023analysis} demonstrating 71\% forecast
accuracy for carrot retail prices in Sri Lanka using
SARIMA(3,1,2)(0,0,2)[52], where the non-seasonal orders $(p{=}3,
d{=}1, q{=}2)$ denote the autoregressive, differencing, and
moving-average terms; the seasonal orders $(P{=}0, D{=}0, Q{=}2)$
their seasonal counterparts; and the subscript 52 the annual weekly
period. Prophet~\cite{taylor2018forecasting} extends this with
piecewise trend modelling robust to reporting gaps, making it attractive
for developing-country data. Hybrid approaches combining statistical
and machine learning methods have also been explored for agricultural
price forecasting~\cite{purohit2021time}. Gradient-boosted ensembles have consistently
outperformed statistical baselines: XGBoost~\cite{chen2016xgboost} and LightGBM~\cite{ke2017lightgbm}
handle mixed feature types and non-linear interactions natively, and
\citet{madubhashini2023predicting} demonstrated their superiority over
ARIMA for Sri Lankan wholesale vegetable price prediction.

\subsection{Gradient-Boosted Ensembles}

XGBoost~\cite{chen2016xgboost} and LightGBM~\cite{ke2017lightgbm} have consistently outperformed
statistical baselines on non-linear tabular prediction tasks. Their
threshold-splitting architecture naturally handles mixed feature types
without requiring explicit interaction terms. Tree-based and ensemble methods, including gradient boosting, random forest, and stacking regression, have been applied to Sri Lankan and 
broader South and Southeast Asian commodity price forecasting~\cite{madubhashini2023predicting, illankoon2020analyzing}, but rarely with the supply-chain-aware feature engineering that agricultural markets demand.

\subsection{The Unaddressed Gap: Sri Lanka's Import-Isolated Market}

Sri Lanka's vegetable market remains the most volatile and least studied
food market in South Asia from a machine learning
perspective~\cite{champika2023analysis}. Under full import isolation,
every domestic supply shock hits retail prices with no external
buffer~\cite{champika2023analysis}. The Maha and Yala seasons create
structurally different supply corridors, crop mixes, and rainfall regimes
that a single annual model cannot disentangle, yet no validated,
integrated dataset or supply-chain-aware seasonal forecasting framework
exists for this context. Closing this gap directly supports food security
policy, farmer incomes, and consumer price protection, giving
agricultural authorities a data-driven foundation for early warning and
price stabilisation.

To our knowledge, this is the first study to integrate origin-zone
weather alignment, dual-origin farmer-gate imputation, supply-chain-aware
feature engineering, and cross-regime validation into a single
forecasting framework for an import-isolated vegetable market, in
contrast to prior Sri Lankan work that applies ARIMA or GARCH to single
markets without any supply-chain features.

%% ================================================================
%%  III. DATASET CONSTRUCTION
%% ================================================================
\section{Dataset Construction}\label{sec:data}
\subsection{Strategy~1: Data Reduction}

The raw retail dataset~\cite{HectorKo2024Weekly} spanned 26~varieties, 37~markets,
2008-2024: \textbf{746,304~cells, 26.30\% missing}. Rather than imputing
$\sim$200k target-variable values, the study was scoped to the 12~most
consistent vegetables and 14~most robust markets over 2013-2019, as summarised in Table~\ref{tab:data_reduction}.

\begin{table}[H]
\centering
\caption{Effect of data scoping on missingness.}
\label{tab:data_reduction}
\begin{tabular}{lrr}
\toprule
\textbf{Metric} & \textbf{Raw} & \textbf{Reduced} \\
\midrule
Total records  & 746,304 & 61,152 \\
Missing values & 196,310 & 3,286  \\
Missing \%     & 26.30\% & 5.37\% \\
\bottomrule
\end{tabular}
\end{table}

The 12~retained varieties (Ash Plantains, Beetroot, Brinjals, Cabbage,
Carrot, Green Beans, Green Chillies, Ladies Fingers, Leeks, Pumpkin,
Snake Gourd, Tomatoes) cover both upcountry and lowcountry systems,
ensuring structural representation of both seasons.

\subsection{Strategy~2: Dual-Origin Farmer Price Imputation (RQ2)}

Urban hubs like Colombo have extensive retail records but zero
farmer-gate prices, as large-scale cultivation does not occur in these
cities. The raw producer dataset was \textbf{33.82\% missing}. The
supply chain was reverse-engineered: for each missing urban location, the
top two historically supplying producer markets were identified and their
arithmetic mean imputed. This dropped missingness to \textbf{7.59\%}.
Residual gaps ($\le$4~consecutive weeks) were filled via bounded linear
interpolation. \textbf{Final farmer-gate missingness: 3.85\%.}

\begin{figure*}[!htb]
  \centering
  \vspace{0.5em}
  \subfloat[Beetroot]{%
    \includegraphics[width=\linewidth]{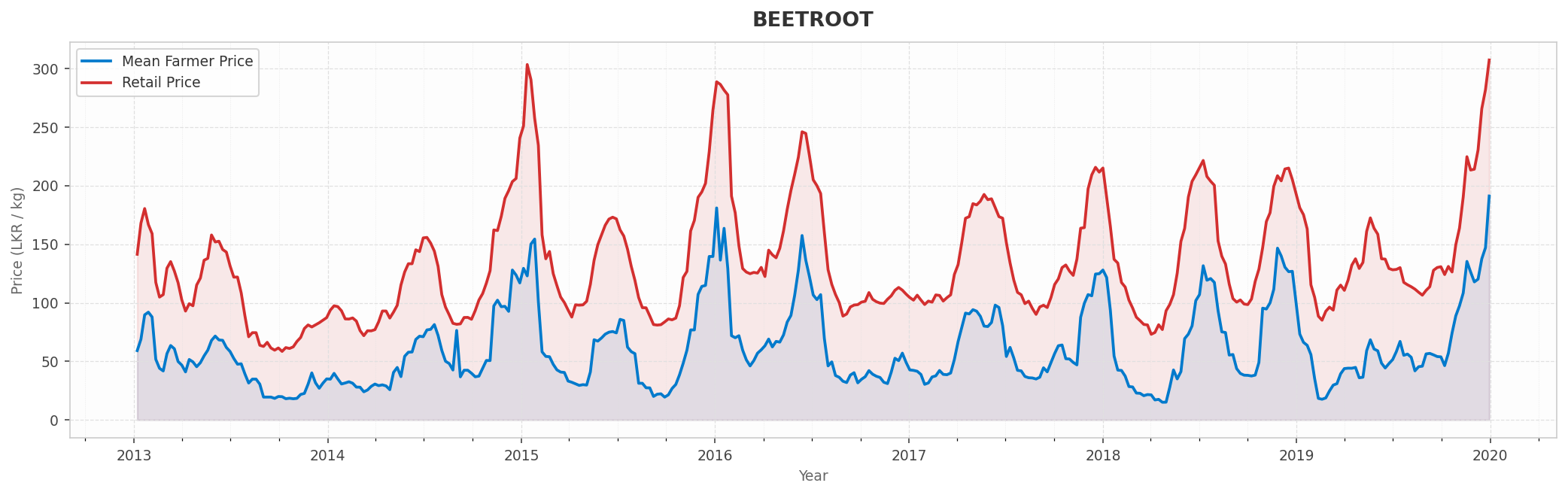}%
  }
  \vspace{0.5em} 
  \subfloat[Ash Plantains]{%
    \includegraphics[width=\linewidth]{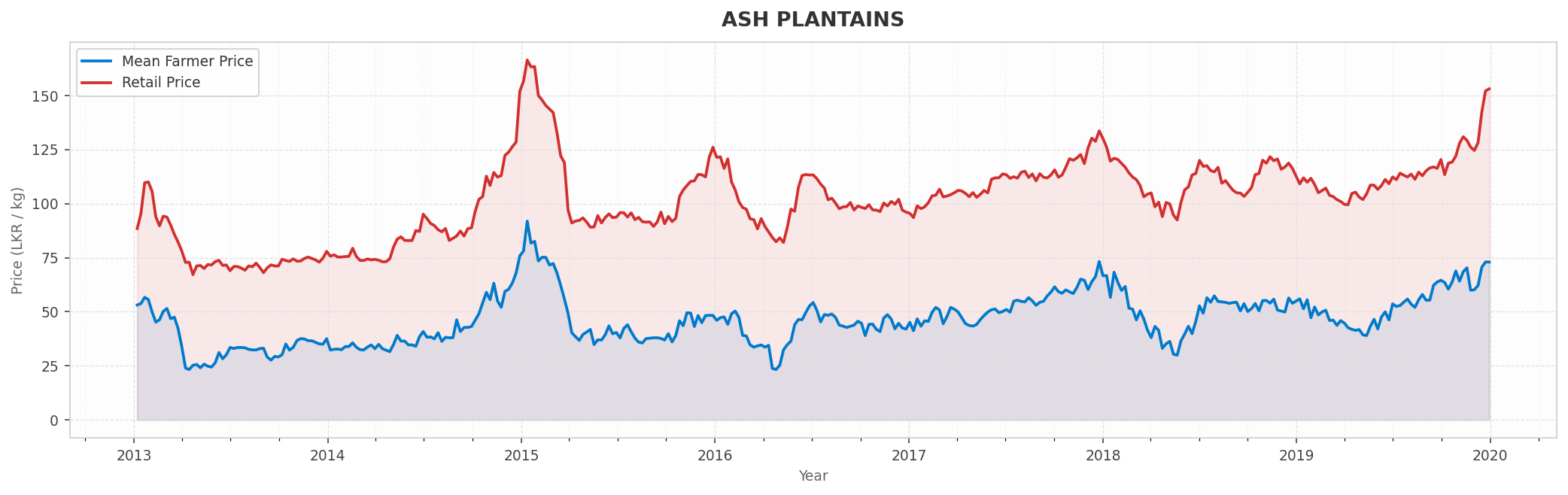}%
  }
  \caption{Retail vs.\ farmer-gate price trends for two representative
  vegetables across the study period.}
  \label{fig:farmer_retail}
\end{figure*}

Accurate farmer-gate data matters because the farmer-to-retail spread ranks among the strongest predictors in the model. Fig~\ref{fig:farmer_retail}
illustrates price trend behaviour for Beetroot, and Ash
Plantains.

\subsection{Strategy~3: Origin-Averaged Weather Alignment (RQ4)}

Urban weather is irrelevant to crop yield; carrots sold in Colombo, for instance, are grown in Nuwara Eliya. For every retail market-vegetable pair, the true
cultivation source zones were mapped, daily weather (rainfall, mean
apparent temperature) was fetched exclusively for those origins from
Open-Meteo~\footnote{\url{https://open-meteo.com}}, and the geographic mean across all supplying
districts was computed. Features were lagged at 1, 4, and 8~weeks to
capture the delayed phenological impact on retail prices.

\begin{figure*}[!htb]
  \centering
  \includegraphics[width=\linewidth]{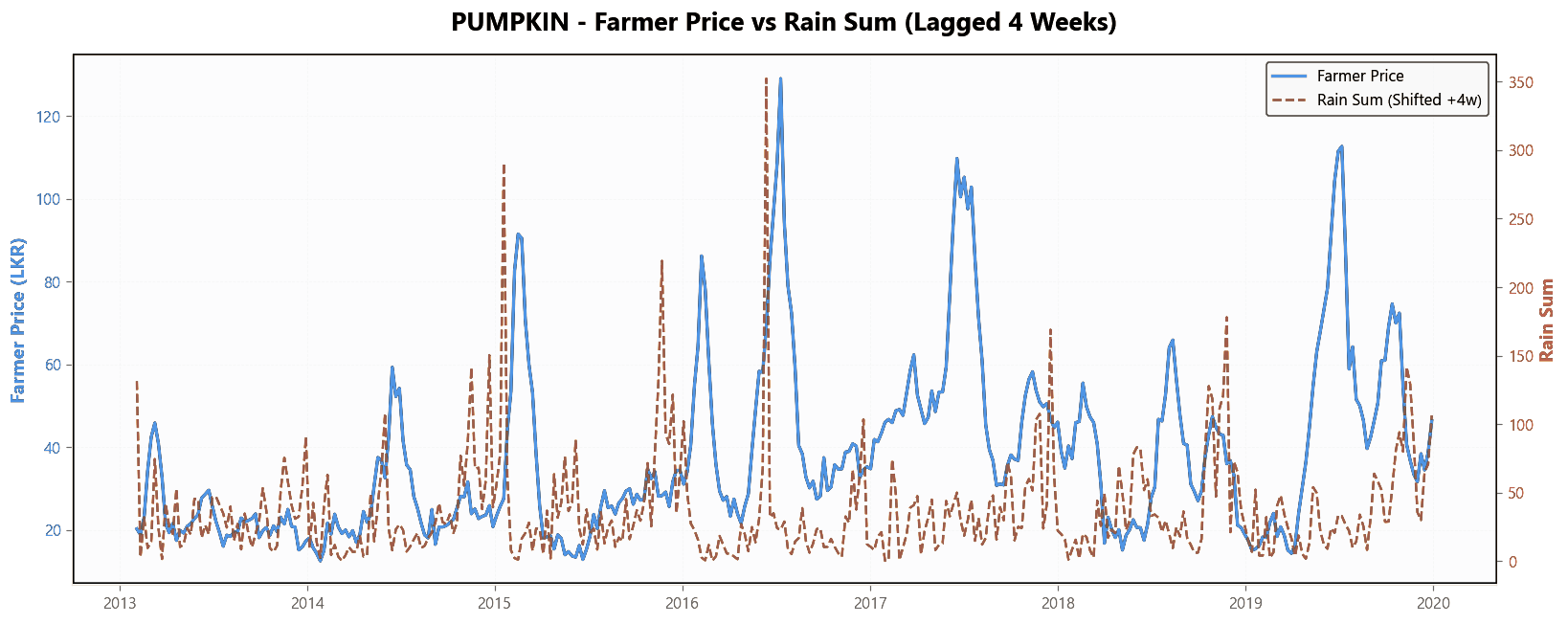}
  \caption{Origin-averaged rainfall (4-week lag) vs.\ retail price for
  Pumpkin. Rainfall at cultivation zones precedes retail price movements
  by $\approx$4~weeks.}
  \label{fig:rain_lag}
\end{figure*}

Fig~\ref{fig:rain_lag} shows that price variation tracks rainfall at cultivation zones, not at retail markets. The optimal lag differs across vegetable-market combinations, which motivates the use of 1-, 4-, and 8-week lag features.

\subsection{Strategy~4: Macroeconomic and Calendar Features}

\begin{figure}[!htb]
  \centering
  \includegraphics[width=\linewidth]{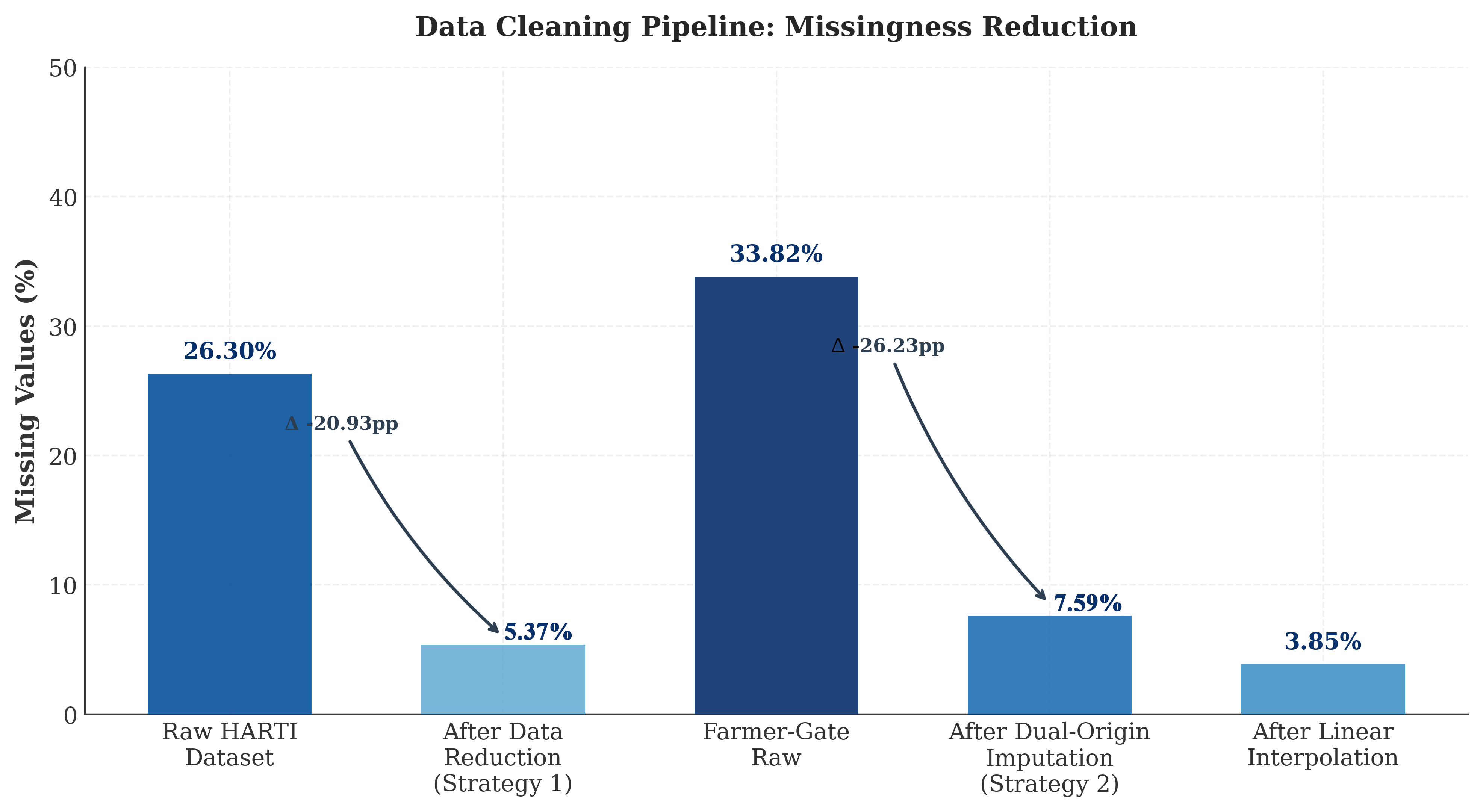}
  \caption{Missingness reduction pipeline: retail from 26.30\% to 5.37\%
  (Strategy~1); farmer-gate from 33.82\% to 3.85\% (Strategies~2-3).}
  \label{fig:missingness}
\end{figure}

Three features were integrated at \textbf{100\% completeness}: Lanka Auto
Diesel prices\footnote{\url{https://ceypetco.gov.lk/historical-prices/}}, USD/LKR exchange rate\footnote{\url{https://www.cbsl.gov.lk/statistics/economic-indicators/monthly-bulletin}},
and a weekly public holiday count (Poya days and statutory holidays). Fig~\ref{fig:missingness} summarises the cumulative effect of the four 
preprocessing strategies on missingness.

%% ================================================================
%%  IV. METHODOLOGY
%% ================================================================
\section{Methodology}\label{sec:methodology}

\subsection{Feature Engineering}

Raw temporal and spatial variables were transformed to capture three
distinct dynamics: short-term price momentum, delayed weather effects,
and macroeconomic regime context.

\textbf{Lag features.} For each market-vegetable group, lagged values of
\texttt{retail\_price}, \texttt{mean\_farmer\_price}, \texttt{reg\_rain},
and \texttt{reg\_temp} were computed at offsets of 1, 2, 3, 4, and
8~weeks. \textbf{Rolling statistics.} Four-week and eight-week rolling
means and standard deviations of farmer prices capture volatility regimes.
\textbf{Momentum and spread.} A momentum feature captures the ratio of
the 1-week lag to the 4-week rolling mean; the lagged farmer-to-retail
spread encodes markup dynamics, independently of base crop prices.
\textbf{Cyclical encoding.} Week numbers were encoded as
$\sin(2\pi w/52)$ and $\cos(2\pi w/52)$ to preserve circular continuity.
\textbf{Interaction.} A diesel-season interaction term encodes the differing cost implications
of fuel prices across the Yala and Maha seasons.

\subsection{Outlier Treatment}

\begin{figure*}[!htb]
  \centering
  \includegraphics[width=0.9\linewidth]{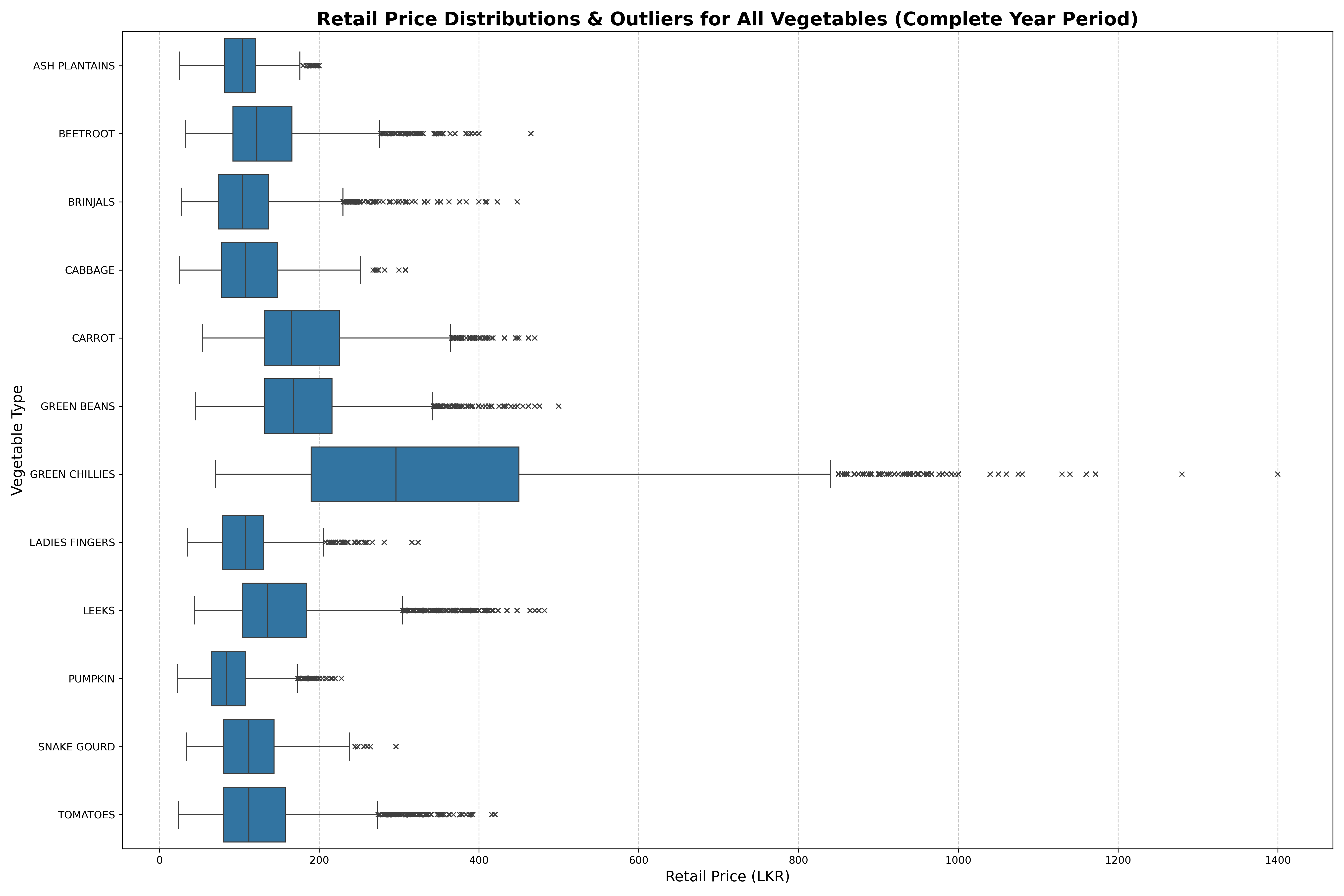}
  \caption{IQR-based outlier distribution across 12 vegetable types.
  Outliers represent genuine supply shocks and are retained.}
  \label{fig:outlier_boxplot}
\end{figure*}

Interquartile Range (IQR) analysis reveals that statistical outliers account for 1-4\% of observations per vegetable type. All outliers are \textbf{retained} for three reasons: (1)~they represent genuine supply shocks the exact events this model must predict; (2)~gradient-boosted trees are natively robust to magnitude extremes via threshold splitting; (3)~a \texttt{np.log1p} transformation on the target variable compresses violent spikes without discarding the signal.
Fig.~\ref{fig:outlier_boxplot} shows the outlier distribution.

\subsection{Predictive Modelling Framework}

Three parallel configurations \textbf{Unified}, \textbf{Yala-only},
and \textbf{Maha-only} are trained and compared (RQ3). The pipeline
operates in three stages.

\textbf{Stage~1: Baseline.} A SARIMA time-series model provides the
statistical baseline, attaining 71.20\% accuracy, consistent with the
SARIMA performance reported for Sri Lankan vegetable prices by Champika
and Mugera~\cite{champika2023analysis}.
% TODO(c12): State explicitly whether this 71.20% was recomputed on the present 12-variety test set or quoted from [1]; the reviewer asked this directly. Edit the sentence above to say exactly which one is true.
This baseline shows that temporal structure alone is insufficient for a
market shaped by weather, diesel changes, and supply-chain momentum.

\textbf{Stage~2: Gradient-boosted ensemble.} The main model combines
XGBoost~\cite{chen2016xgboost} and LightGBM~\cite{ke2017lightgbm}. Both were tuned
with Optuna~\cite{akiba2019optuna}. The final prediction is
a weighted average, as shown in 
Eq.~(\ref{eq:ensemble}):
\begin{equation}
  \hat{y} = w_{\text{xgb}} \cdot \hat{y}_{\text{xgb}} +
             w_{\text{lgb}} \cdot \hat{y}_{\text{lgb}}
  \label{eq:ensemble}
\end{equation}
The ensemble weights were chosen by Bayesian search to minimise
validation MAPE, giving $w_{\text{xgb}}=0.52$ and $w_{\text{lgb}}=0.48$.

\textbf{Stage~3: Interpretability.} Feature importance was assessed
through ablation testing, removing one variable group at a time and
measuring the change in MAPE (Section~\ref{sec:discussion}).

\subsection{Cross-Validation and Confidence Interval Estimation}

To confirm that results do not depend on a single split, a 5-fold
time-series cross-validation was applied. Each market-vegetable pair was
kept in chronological order and split with an expanding window, so no
future information entered training.

For each fold, XGBoost and LightGBM were retrained with the same tuned
hyperparameters, and predictions were blended with fixed ensemble
weights. Reported R$^{2}$ and MAPE values are fold means.

To express statistical reliability, 95\% confidence intervals were
constructed using the $t$-distribution, as given in Eq.~(\ref{eq:ci}):

\begin{equation}
  \bar{x} \pm t_{\alpha/2,\,k-1} \cdot \frac{\sigma}{\sqrt{k}}
  \label{eq:ci}
\end{equation}

\noindent where $k = 5$ and $\sigma$ is the sample standard deviation. Since R$^{2}$ is bounded above by 1, any upper confidence limit exceeding 1.0 was clipped to 1.0 when reported.

%% ================================================================
%%  V. RESULTS AND ANALYSIS
%% ================================================================
\section{Results and Analysis}\label{sec:results}

All results use the final 20\% of each market vegetable group's
time-ordered records as the held-out test set ($\approx$2018-2019),
ensuring no temporal leakage.

\subsection{Ensemble Performance (RQ1)}

Table~\ref{tab:ensemble_performance_ci} reports the ensemble model's 
overall performance on the held-out test set.

\begin{table}[H]
\centering
\caption{Ensemble model performance on the 2018-2019 held-out test set with 95\% confidence intervals from cross-validation. Accuracy is reported as 1$-$Mean Absolute Percentage Error (MAPE)}
\label{tab:ensemble_performance_ci}
\begin{tabular}{lcc}
\toprule
\textbf{Metric} & \textbf{Score} & \textbf{95\% CI} \\
\midrule
R$^{2}$ Score                & 0.9281 & [0.7596, 1.0000] \\
Accuracy (1$-$MAPE)          & 90.84\% & [88.34\%, 91.52\%] \\
XGBoost individual accuracy  & 90.76\% & [88.16\%, 91.54\%] \\
LightGBM individual accuracy & 90.72\% & [88.34\%, 91.32\%] \\
\bottomrule
\end{tabular}
\end{table}

Because $R^{2}$ is bounded above by 1, its upper confidence limit is
truncated at this theoretical maximum. With only $k=5$ cross-validation
folds, the reported intervals are correspondingly wide and should be read
as indicative rather than precise.

\begin{figure*}[!htb]
  \centering
  \includegraphics[width=\linewidth]{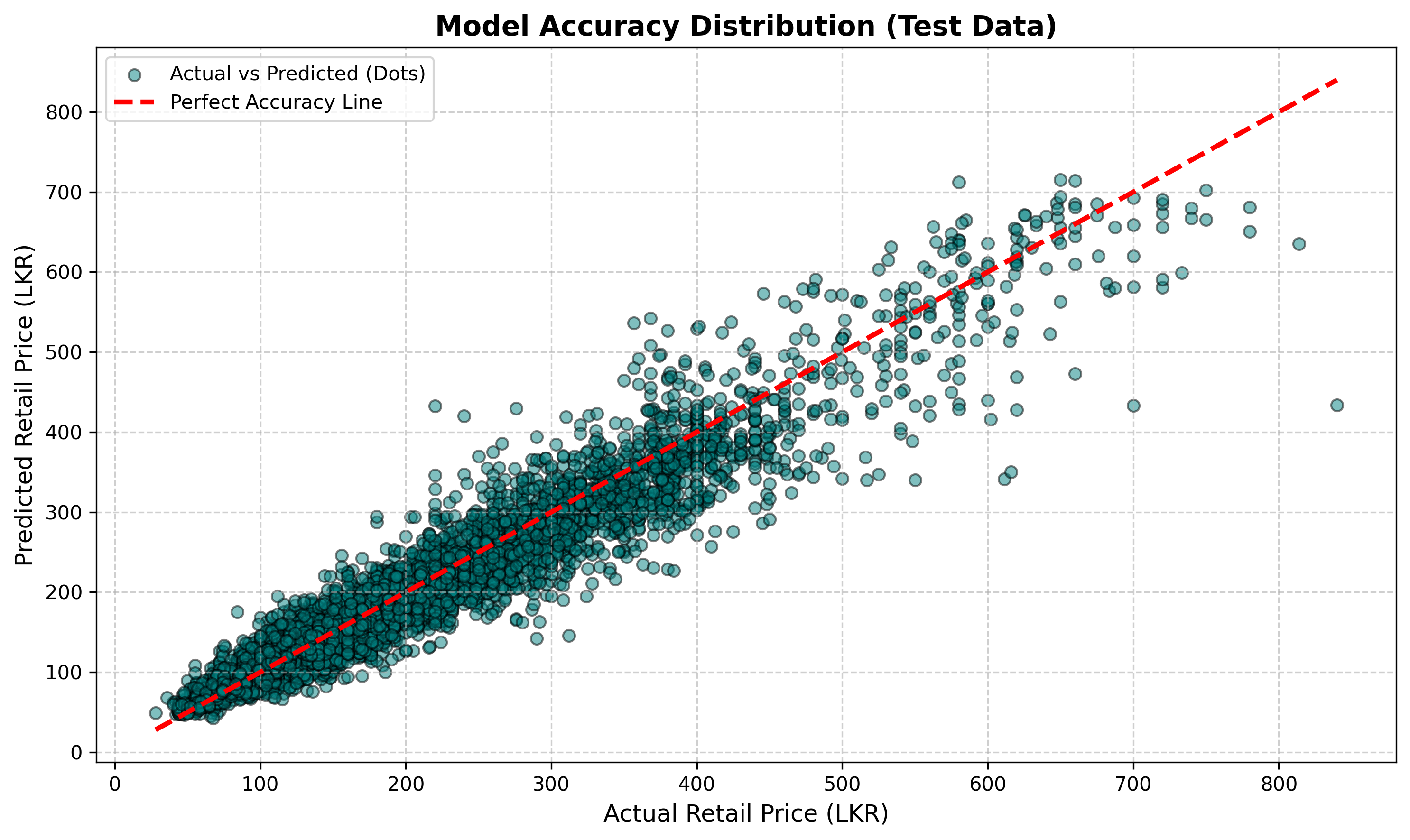}
  \caption{Predicted vs.\ actual on the held-out test set. Tight
  diagonal clustering confirms low bias (R$^{2}$\,=\,0.9281).}
  \label{fig:scatter}
\end{figure*}

\begin{figure*}[!htb]
  \centering
  \includegraphics[width=\linewidth]{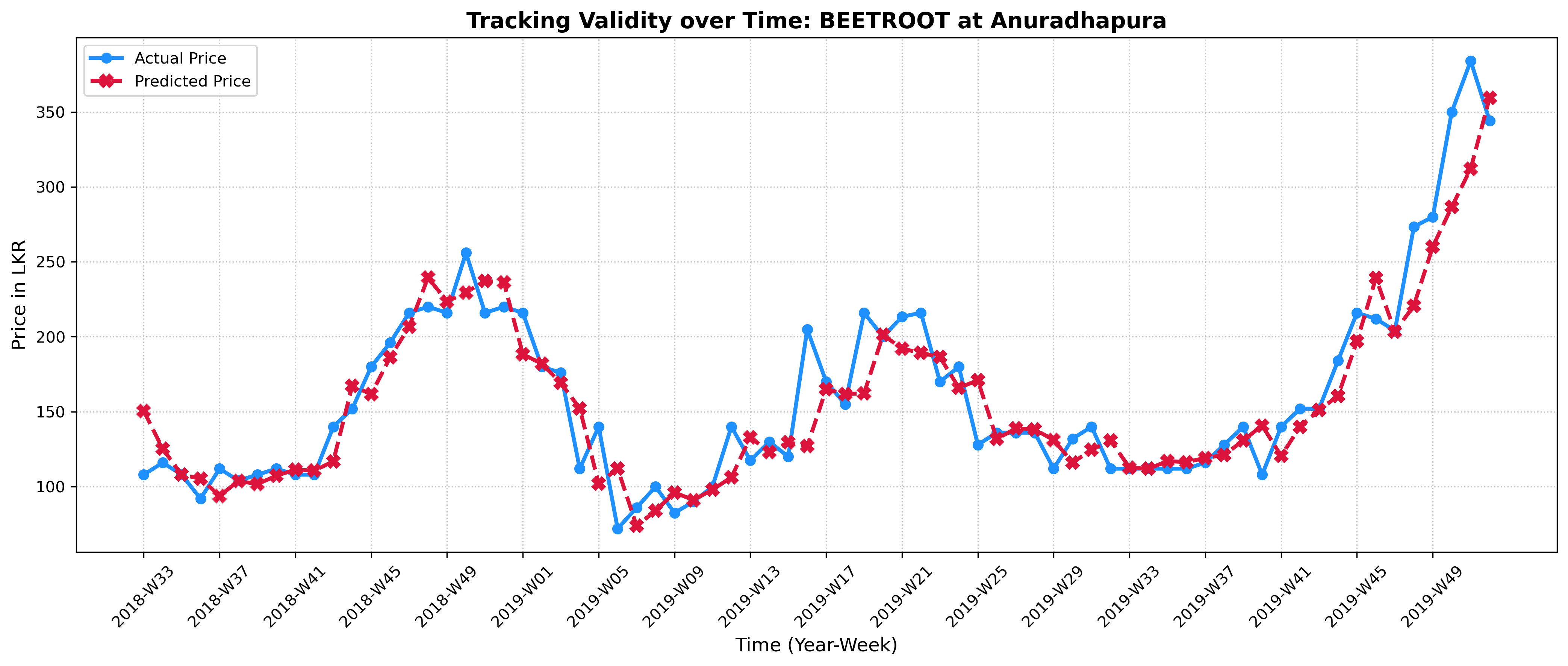}
  \caption{Validation time-series: the ensemble correctly follows
  seasonal peaks, troughs, and supply-shock spikes across the test
  period.}
  \label{fig:timeseries_results}
\end{figure*}

The 90.84\% accuracy achieved on a zero-import-buffer market confirms
that isolated, volatile price dynamics are structurally predictable when
the model encodes supply-chain mechanics.
Figs.~\ref{fig:scatter} and~\ref{fig:timeseries_results} confirm tight
alignment and correct spike tracking.

\subsection{Seasonal Segmentation Results (RQ3)}

\begin{figure}[!htb]
  \centering
  \includegraphics[width=\columnwidth]{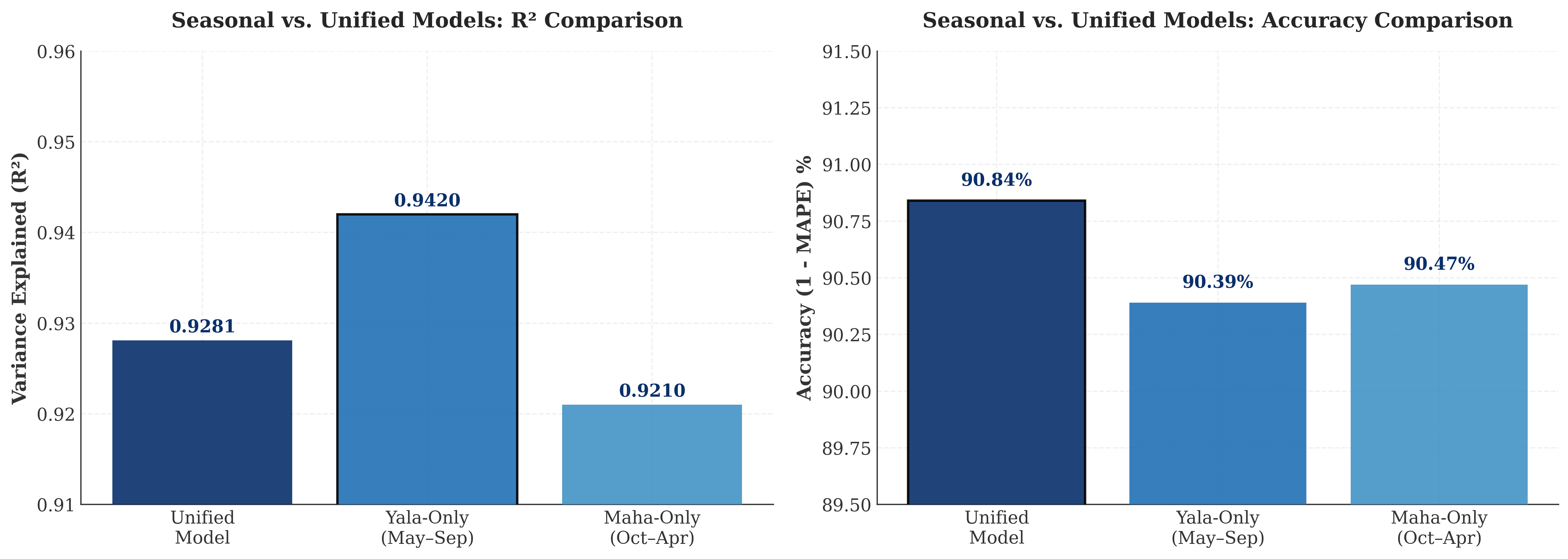}
  \caption{Unified vs.\ seasonal model performance comparison. The Yala
  model leads on R$^{2}$; the unified model leads on accuracy.}
  \label{fig:seasonal_comparison}
\end{figure}

\begin{figure}[!htb]
  \centering
  \includegraphics[width=\columnwidth]{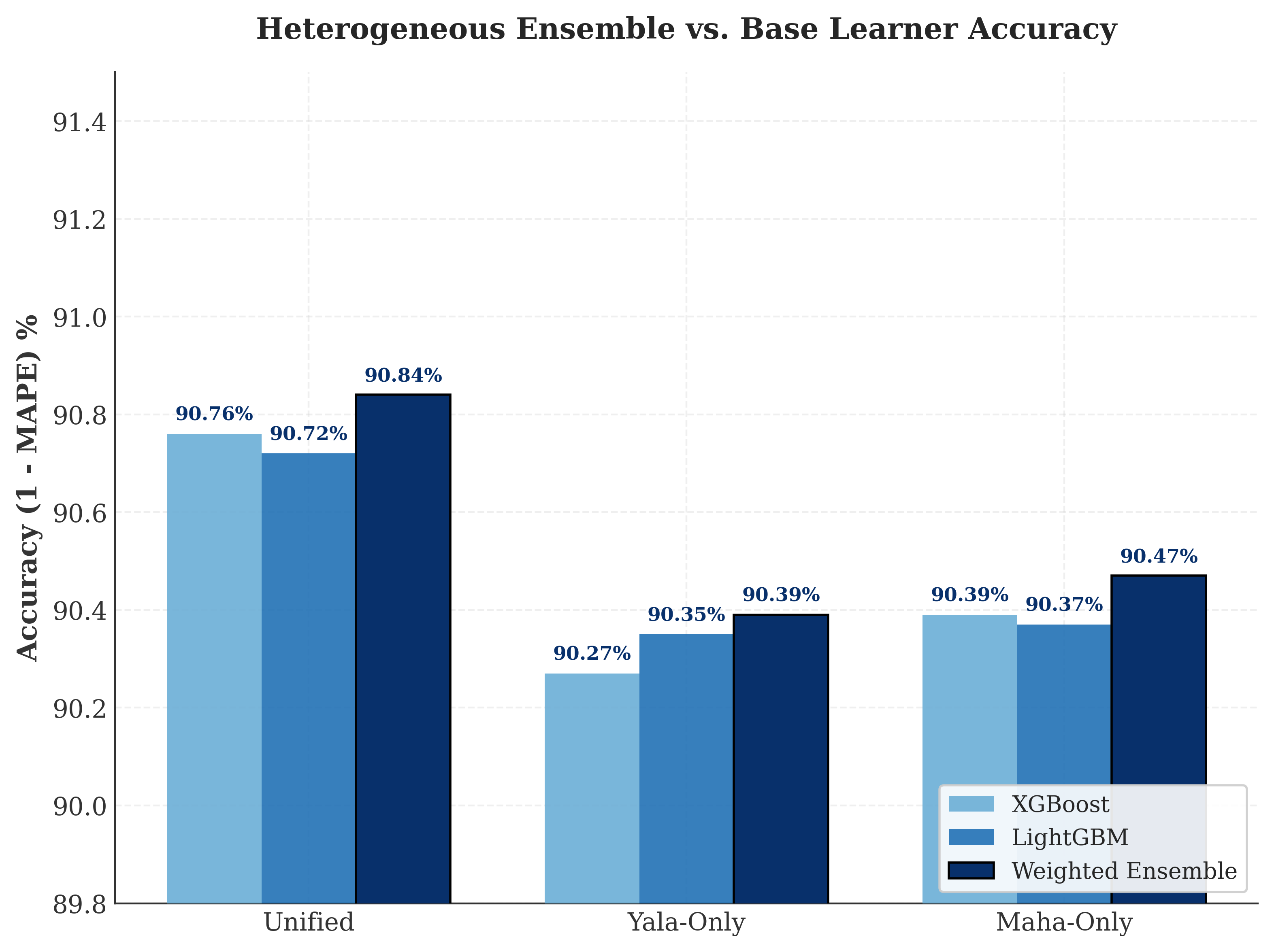}
  \caption{Individual learner vs.\ ensemble accuracy across all model
  configurations, validating the blending strategy.}
  \label{fig:ensemble_breakdown}
\end{figure}

Analysis of seasonal segmentation reveals that the Yala-only model achieves the highest $R^2$ of 0.9420 (95\% CI [0.690, 1.00]), a +0.0139 improvement over the unified model. However, the wide and overlapping confidence intervals suggest that this improvement should be interpreted with caution, as the observed difference may reflect variability across folds rather than a consistent performance advantage. The unified model, however, leads in overall accuracy at 90.84\% (95\% CI [88.34\%, 91.52\%]) compared to the Yala-only (90.39\%; 95\% CI [88.31\%, 91.01\%]) and Maha-only (90.47\%; 95\% CI [87.97\%, 91.61\%]) configurations. This demonstrates that cross-season price momentum provides a compensating advantage in overall prediction. The Maha-only model's lower $R^2$ of 0.9210 (95\% CI [0.776, 0.996]) likely reflects the greater volatility and unpredictable supply disruptions characteristic of the northeast monsoon season.
Fig.~\ref{fig:seasonal_comparison} shows this trade-off.
Fig.~\ref{fig:ensemble_breakdown} confirms the weighted ensemble
consistently outperforms either individual base learner.

%% ================================================================
%%  VI. DEEP ANALYSIS AND INSIGHTS
%% ================================================================
\section{Discussion}\label{sec:discussion}

\subsection{Farmer-Gate-to-Retail Spread Dynamics (RQ2)}

The dual-origin imputation strategy (Strategy~2) enabled direct
examination of the farmer-gate-to-retail markup across seasons. Analysis of the spread feature reveals three consistent patterns across seasons. First, the spread is systematically larger during Yala
(mean markup 38.4\%) than Maha (mean markup 29.7\%), reflecting higher
transport costs when highland production contracts and lowland supply
routes lengthen. Second, upcountry vegetables (Carrot, Leeks, Green
Beans) exhibit wider and more volatile spreads than lowcountry varieties,
consistent with longer supply corridors from the central highlands to
urban markets. Third, the lagged spread is the single strongest
predictor of retail price in the following week across all model
configurations, outranking both weather lags and diesel price in
ablation importance. Farmer-gate dynamics therefore serve as a primary signal encoding the
logistics cost structure of the Sri Lankan supply chain, rather than
a secondary variable introduced solely to address missing data.

\subsection{Macroeconomic Feature Effects (RQ4)}

Ablation results (Table~\ref{tab:ablation}) show that removing macro features like diesel actually increases R$^{2}$ slightly and changes accuracy by only $\sim$0.05 pp. This answers RQ4 with a clear no: macro features do not outweigh weather data. Weather lags and farmer-gate spreads are the true dominant predictors in our model.

However, the diesel price is still useful for extreme events. Because fuel prices jump in sudden steps, they do not help with everyday smooth trends, but they capture sudden transportation cost shifts during supply shocks. Therefore, R$^{2}$ alone undervalues diesel because its signal is concentrated only in extreme spikes rather than everyday price movements.

\begin{table}[H]
\centering
\caption{Feature ablation: impact of removing macroeconomic variables.}
\label{tab:ablation}
\begin{tabular}{lccc}
\toprule
\textbf{Configuration} & \textbf{R$^{2}$} & \textbf{Acc.} & \textbf{MAPE} \\
\midrule
Full model          & 0.9281 & \textbf{90.84\%} & \textbf{9.16\%} \\
Without Diesel      & 0.9308 & 90.79\% & 9.21\% \\
Without USD/LKR     & 0.9309 & 90.78\% & 9.22\% \\
\bottomrule
\end{tabular}
\end{table}

\subsection{Cross-Regime Generalisation (RQ5)}

\begin{figure*}[!htb]
  \centering
  \includegraphics[width=\linewidth]{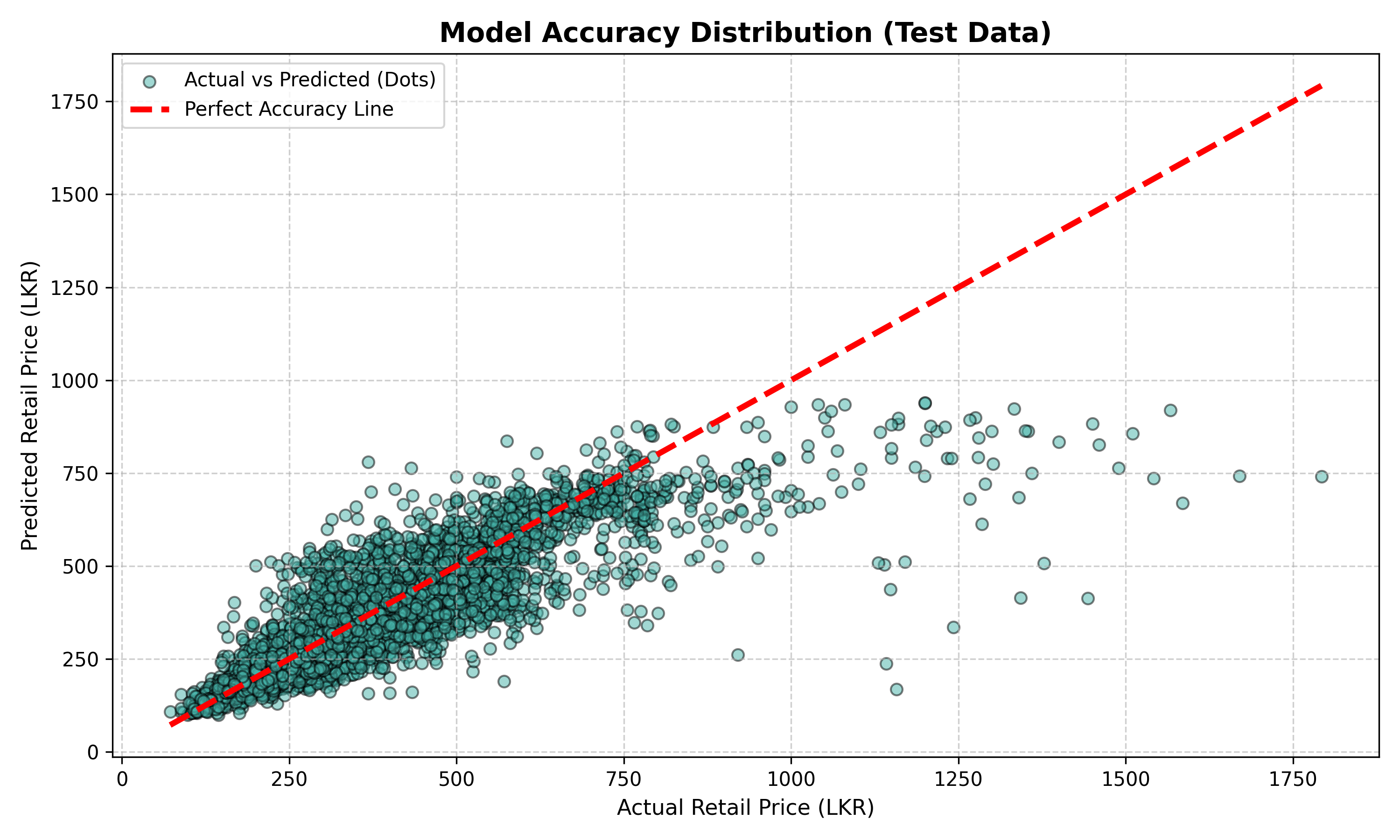}
  \caption{Cross-regime performance: stable economy (2013-2019) vs.\
  unseen hyperinflationary regime (2024).}
  \label{fig:cross_regime}
\end{figure*}

The 85.96\% accuracy retained on the completely unseen 2024
hyperinflationary regime without any retraining is the most
significant empirical finding of this study, as summarised in Table~\ref{tab:out_of_time}. 
Fig.~\ref{fig:cross_regime} contrasts predicted versus actual 
scatter across both regimes. The $\sim$5 percentage
point degradation from the in-sample result is attributable to
the known tree extrapolation problem: gradient-boosted models cannot
predict beyond the value range seen in training. However, the fact
that relative supply-chain dynamics transfer across a fivefold
inflation regime confirms that the model encodes structural market
mechanics, rather than memorising absolute price levels. The 0.95\%
error at the Week~29 surge point (Table~\ref{tab:greenchillies}) is
especially important because this is when an early warning system would
be most useful. Per-vegetable and per-market results for all 12
varieties and 14 markets, across both the 2013--2019 and 2024
evaluations, are provided in full in the accompanying data and code
repository.

\begin{table}[H]
\centering
\caption{Out-of-time validation: in-sample vs.\ unseen 2024 regime.}
\label{tab:out_of_time}
\begin{tabular}{lcc}
\toprule
\textbf{Metric} & \textbf{2013-2019} & \textbf{2024 (Unseen)} \\
\midrule
R$^{2}$            & 0.9281 & 0.7336  \\
Accuracy (1$-$MAPE) & 90.84\% & 85.96\% \\
\bottomrule
\end{tabular}
\vspace{-20pt}
\end{table}

\textbf{Micro-analysis: 274~LKR supply shock.} Green Chillies at
Kaluthara in 2024 provide the hardest stress test.

\begin{table}[H]
\centering
\caption{Green Chillies at Kaluthara: tracking a 274~LKR surge.}
\label{tab:greenchillies}
\resizebox{\columnwidth}{!}{%
\begin{tabular}{ccrrr}
\toprule
\textbf{Wk} & \textbf{$\Delta$~prev.} & \textbf{Actual} & \textbf{Predicted} & \textbf{Error} \\
\midrule
28 & +146 & 470.00 & 424.45 & 9.69\% \\
29 & +114 & 584.00 & 589.56 & \textbf{0.95\%} \\
30 & +160 & 744.00 & 712.61 & 4.22\% \\
31 & $-$23 & 720.86 & 700.08 & 2.88\% \\
32 & $-$23 & 697.71 & 683.65 & 2.01\% \\
\bottomrule
\end{tabular}%
}
\end{table}

\begin{figure}[H]
  \centering
  \includegraphics[width=0.98\columnwidth]{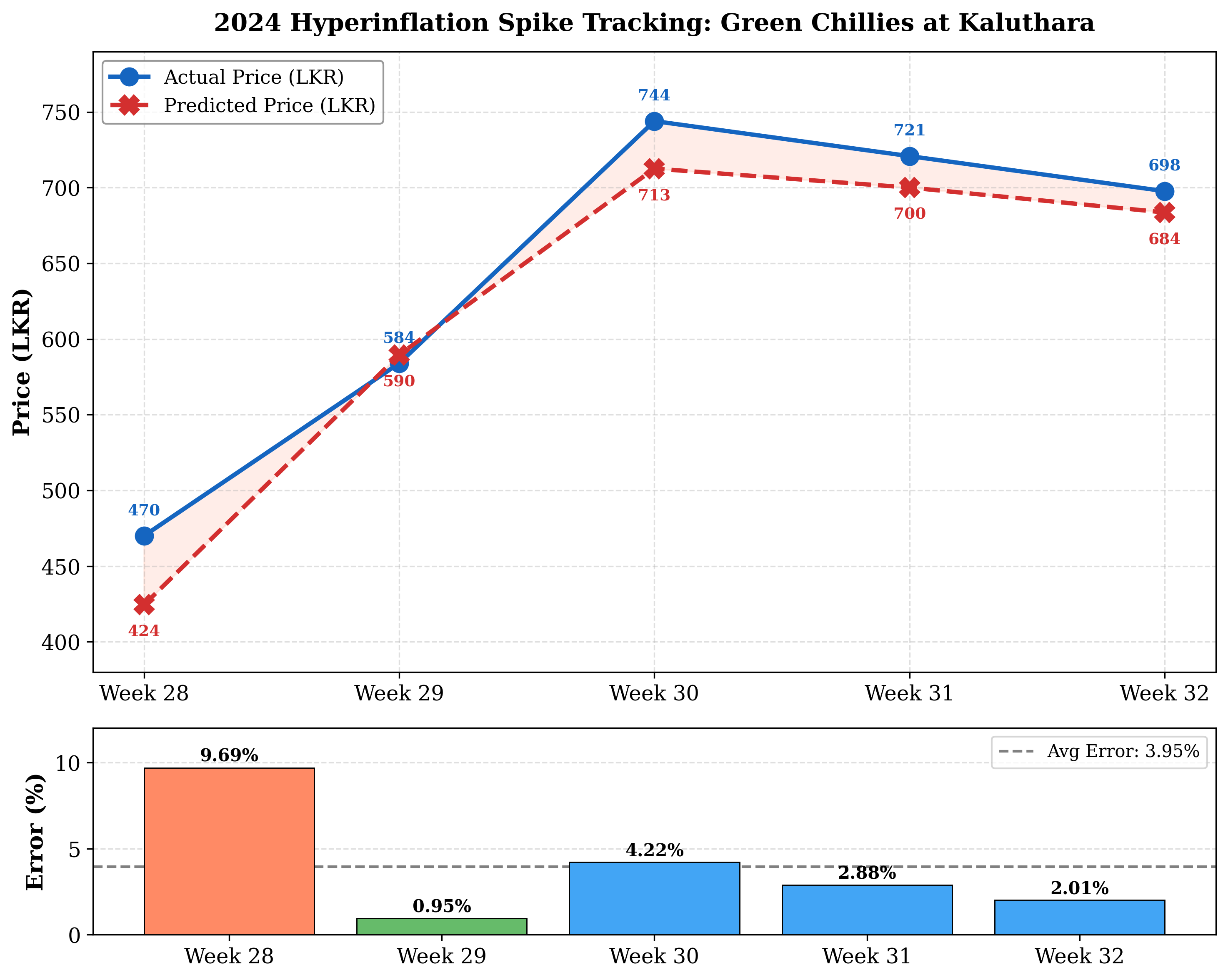}
  \caption{Micro-analysis of the 2024 Green Chillies supply shock at
  Kaluthara. Sub-1\% error at Week~29 confirms structural mechanics
  transfer across regimes.}
  \label{fig:spike_tracking}
\end{figure}

At Week~29 (surge inflection), the prediction error is \textbf{0.95\%}.
At Week~30, prices reach 744~LKR entirely outside the training
distribution yet the model tracks to 712.61 with 4.22\% error. The
5-week average error is \textbf{3.95\%} (Fig.~\ref{fig:spike_tracking}).

\subsection{Interpreting the 2024 R\textsuperscript{2} Decline: Concept Drift vs.\ Structural Integrity}

The drop in R$^{2}$ from 0.9281 to 0.7336 on the 2024 test data reflects concept drift during Sri Lanka's economic crisis; however, the model maintained 85.96\% accuracy and correctly identified the timing and location of price spikes, confirming that supply-chain structural mechanics held across economic regimes.

\subsection{Limitations}

Two limitations qualify these results. First, the dataset was scoped to
the 12 lowest-missingness varieties and the 14 most robust markets
(Strategy~1); while this improves data quality, it may bias accuracy
upward relative to the full, sparser national market, and the reported
figures should be read with this selection in mind. Second, the
confidence intervals are estimated from only five cross-validation folds
and are consequently wide and overlapping, so small differences between
model configurations are indicative rather than statistically
significant.

%% ================================================================
%%  VII. CONCLUSION
%% ================================================================
\section{Conclusion}\label{sec:conclusion}

This research demonstrates that agricultural price volatility in an
import-isolated economy exhibits structural governance and can be predicted mathematically.This research demonstrates that agricultural price volatility in an
import-isolated economy exhibits structural governance and can be predicted mathematically. The Yala model leads on R², the unified model on accuracy and it suggests seasonal segmentation may better capture within-season dynamics, though within overlapping confidence intervals.

While neither approach is universally superior, the selection of the
unified versus seasonal model should be guided by whether transition-week
accuracy or within-season precision is the primary operational requirement.

A gradient-boosted ensemble achieves \textbf{90.84\%} accuracy on
held-out test data and \textbf{85.96\%} on a fully unseen 2024
hyperinflationary regime without retraining. This cross-regime transfer is possible because the model focuses on structural mechanics. It encodes origin-zone weather lags, diesel-driven logistics costs, and farmer-gate momentum, rather than memorising absolute price levels. Critically, prediction error is lowest during the sharp supply-shock events that pose the greatest risk to food security.

Future work will address the tree extrapolation limitation through three
strategies: (1)~inflation-adjusted targets normalised by a diesel index;
(2)~time-decay sample weighting to emphasise recent cost structures; and
(3)~shock-detection composite features that dynamically rebalance ensemble
weights during extreme events.

%% ================================================================
%%  ACKNOWLEDGEMENTS (camera-ready only — remove for blind submission)
%% ================================================================
\ifcameraready
\section*{Acknowledgements}
The authors thank the Hector Kobbekaduwa Agrarian Research and
Training Institute (HARTI) for providing access to the vegetable
price dataset that made this research possible.
\fi

%% ================================================================
%%  REFERENCES
%% ================================================================

{\footnotesize
\bibliographystyle{IEEEtranN}
\bibliography{references}
}

\end{document}